\documentclass[conference,compsoc]{IEEEtran}
 
\usepackage{amsfonts}

\usepackage{subfig}
\usepackage{booktabs} %

\usepackage{dsfont}

\usepackage{amssymb}
\usepackage{mathtools}
\usepackage{amsthm}
\usepackage{mathabx}
\usepackage{breqn}

\usepackage{algorithm}
\usepackage{algpseudocode}
\usepackage{float}

\DeclareMathOperator*{\argmax}{arg\,max}

\newcommand{\norm}[1]{\left\lVert#1\right\rVert}

\usepackage{xspace}
\newcommand{\etal}[0]{et al.\xspace} %
\newcommand\floor[1]{\lfloor#1\rfloor}

\newcommand{\eat}[1]{}

\usepackage{hyperref}

\usepackage{amsmath}
\usepackage{amssymb}
\usepackage{mathtools}
\usepackage{amsthm}

\usepackage[capitalize,noabbrev]{cleveref}

\theoremstyle{plain}
\newtheorem{theorem}{Theorem}[section]

\newtheorem{lemma}[theorem]{Lemma}
\newtheorem{corollary}[theorem]{Corollary}
\theoremstyle{definition}
\newtheorem{definition}[theorem]{Definition}

\theoremstyle{remark}

\newcommand{\reals}{\mathbb{R}}

\usepackage{tikz}
\newcommand\copyrighttext{%
  \footnotesize \textcopyright Accepted for publication at IEEE Security \& Privacy, 2024.}
\newcommand\copyrightnotice{%
\begin{tikzpicture}[remember picture,overlay]
\node[anchor=south,yshift=10pt] at (current page.south) {\fbox{\parbox{\dimexpr\textwidth-\fboxsep-\fboxrule\relax}{\copyrighttext}}};
\end{tikzpicture}%
}

\begin{document}

\title{It's Simplex! Disaggregating Measures to Improve Certified Robustness
\thanks{Identify applicable funding agency here. If none, delete this.}
}

    \author{\IEEEauthorblockN{Andrew C. Cullen\IEEEauthorrefmark{1}\IEEEauthorrefmark{3},
            Paul Montague\IEEEauthorrefmark{2},
            Shijie Liu\IEEEauthorrefmark{1}, 
            Sarah M. Erfani\IEEEauthorrefmark{1}, and
            Benjamin I.P. Rubinstein \IEEEauthorrefmark{1},~\IEEEmembership{Fellow,~IEEE}}
        \IEEEauthorblockA{\IEEEauthorrefmark{1}School of Computing and Information Systems,
            University of Melbourne, Parkville, Australia}
        \IEEEauthorblockA{\IEEEauthorrefmark{2}Defence Science and Technology Group, Adelaide, Australia}
        \IEEEauthorblockA{\IEEEauthorrefmark{3} andrew.cullen@unimelb.edu.au}%
        }

\maketitle
\copyrightnotice

\begin{abstract}

Certified robustness circumvents the fragility of defences against adversarial attacks, by endowing model predictions with guarantees of class invariance for attacks up to a calculated size. While there is value in these certifications, the techniques through which we assess their performance do not present a proper accounting of their strengths and weaknesses, as their analysis has eschewed consideration of performance over individual samples in favour of aggregated measures. By considering the potential output space of certified models, this work presents two distinct approaches to improve the analysis of certification mechanisms, that allow for both dataset-independent and dataset-dependent measures of certification performance. Embracing such a perspective uncovers new certification approaches, which have the potential to more than double the achievable radius of certification, relative to current state-of-the-art. Empirical evaluation verifies that our new approach can certify $9\%$ more samples at noise scale $\sigma = 1$, with greater relative improvements observed as the difficulty of the predictive task increases.

\end{abstract}


\section{Introduction}
\label{sec:intro}

Despite their excellent benchmark performance, the black-box nature of deep neural networks makes them prone to unexpected behaviours and instability. Of particular interest are \emph{adversarial examples}, in which model outputs are changed by way of human imperceptible input perturbations~\cite{biggio2013evasion,szegedy2013intriguing,goodfellow2014explaining, cullen2023paradox}. The spectre of such attacks poses risks to deployed models, in a fashion that can materially impact both the model deployer and their users.

While numerous defences against the mechanisms that produce these examples exist, they typically only tackle a single vulnerability and be circumvented by considering alternate attack mechanisms. This intrinsic limitation motivated the development of \emph{certifiably robust} models, which provide a pointwise guarantee of resistance to attacks up to a fixed, calculable size. These certifications are typically constructed by exploiting either \emph{convex relaxation} or \emph{randomised smoothing}~\cite{lecuyer2019certified}, and measure how close the nearest-possible adversarial example could be, independent of the technique employed to identify said attack.

Whenever a new certification mechanism is proposed, its utility is demonstrated by considering its average performance over a large number of samples. However, these aggregate measures do not align with the motivations of potential attackers, who may seek to attack individual samples. As such, aggregate measures of certification may disguise how the risk profile of individual samples may change, and more broadly removes the ability to interrogate the factors that drive differential performance of different certification schemes.

In contrast to these aggregated measures of performance, this work takes the contrary, disaggregated perspective, and considers how the performance of different certification schemes depends upon where an individual sample exists within the simplex of permissible output scores. This is made possible by way of some simple-yet-powerful observations relating to the analytic nature of certification mechanisms, which allow for analytic, dataset independent comparisons to be performed. These techniques are then merged with a dataset-dependent, sample-wise analysis, which considers a dataset's distribution in the context of the output simplex.

In taking this approach, we are able to both better understand the relative performance of certification schemes, and the nature of adversarial risk in certified systems more broadly. This is of critical importance for deployed systems, in which samples will likely be considered to hold differing levels of adversarial risk. Moreover, it may well be that the distribution properties of standard tested datasets may not reflect those observed within vulnerable deployed systems. If this was true, then being able to assess certification performance in a fashion that is aware but not beholden to the output distribution has significant ramifications for understanding the generalisation of certification techniques. 

This disaggregated perspective reveals the potential for two additions to the certification oeuvre, which we document within this work. The first of which involves a revised mechanism for constructing certifications by way of differential privacy, which for a subset of samples can produce a more than two-fold increase in the achievable certification for what we will categorise as multinomial style certifications, and a more than five-fold increase for softmax certifications. Our second contribution involves treating certifications not as the product of any one mechanism, but rather as the best calculated value across a set of different approaches, an approach that is only made possible by considering certification performance through a disaggregated lens.

These improvements in both how we consider certifications, and how we construct certifications are supported by Sections~\ref{sec:related} and \ref{sec:dataindpt}, which introduce the current range of extant certification approaches, and how their analytic nature can be used to construct dataset-independent comparisons. From this, Section~\ref{sec:improved_mechanism} then considers how these measures can be used to help reveal the potential for our two new certification approaches, both of which have the potential to help improve the sample-wise performance of certifications through consideration of the simplex of permissible output scores. Section~\ref{sec:results} then demonstrates how our new approaches uniformly outperforms prior techniques when considering expectations over models that output softmax probability distributions, while providing significant advantages in a subset of the certification domain for multinomial classifiers.

\begin{figure*}[t]
\begin{center}
\includegraphics[width=0.5\linewidth]{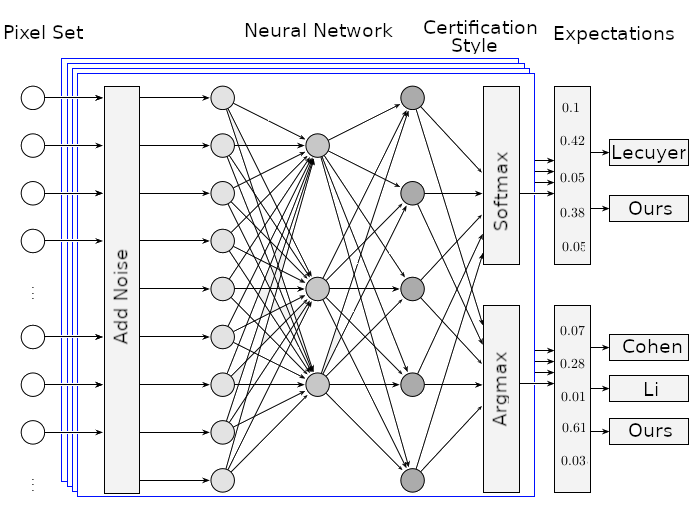}
\end{center}%
\caption{A representation of the process of certifying a single image. Within this diagram, the blue squares represent repeated independent calculations, over which the ensemble expectations are calculated. In this paper, the ensemble itself is taken over the maximum certifications of the Cohen \etal~\cite{cohen2019certified}, Li \etal~\cite{li2018certified}, and Our approaches.}
\label{fig:CVPR}
\end{figure*}

\section{Preliminaries}
\label{sec:related}

Certification mechanisms use a mixture of computational and analytical techniques to provide guarantees of a models resistance to \emph{all} adversarial attack. While this approach can be applied to training-time processes~\cite{liu2023enhancing}, within this work we are specifically interested in guarding against $\ell_2$-norm perturbations to images at evaluation-time.
A learned classifier $f_{\boldsymbol{\theta}} \in \reals^K$ acting upon an input sample $\mathbf{x} \in \reals^{d}$ is considered to be \emph{robust} to attacks $\boldsymbol{\gamma}\in\reals^d$ of bounded size $\|\boldsymbol{\gamma}\|_p\leq L$---henceforth referred to as $B_P(L)$---if
\begin{align}\label{eqn:attack_condition}
    & \argmax_{\forall i \in \mathcal{K}} f_{\boldsymbol{\theta_i}} = j \text{ and } \nonumber \\
    \forall \boldsymbol{\gamma} \in B_p(L)\centerdot &f_{\boldsymbol{\theta_j}} (\mathbf{x}+\boldsymbol{\gamma}) > \max_{i: i\neq j} f_{\boldsymbol{\theta_i}} (\mathbf{x}+\boldsymbol{\gamma})\enspace.\\ 
    & \text{where } \mathcal{K} = \{1, \ldots, K\} \nonumber
\end{align}
The simplicity of this statement stands in stark contrast to the difficulty of proving it, as exploring the entire feasible space of $\boldsymbol{\gamma}$ is computationally intractable, especially as $d$ increases. As such, in order to establish the robustness of a model, certifications mechanisms instead construct \emph{provable lower bounds} on the distance to the nearest adversarial example, making such certifications inherently conservative.

In attempting to certify against $\ell_2$-norm bounded perturbations, two primary frameworks have been considered, which can be broadly categorised as \emph{statistical certifications}, and those that \emph{exploit knowledge of the model's architecture}. Of these, the latter involves constructing bounds on the output of a model by inspecting and tracing bifurcations under norm-bounded perturbations \cite{mirman2018differentiable, weng2018towards, zhang2018efficient, zhang2018efficient, singh2019abstract}. Framed in general as convex relaxation, these techniques opt to use linear relaxation to construct bounding polytopes of a model's outputs over bounded perturbations~\cite{salman2019convex}. These approaches have been extended by adopting augmented loss functions to promote tight output bounds \cite{xu2020automatic}. However, these approaches require significant amounts of computational resources to construct their certifications, which typically leads to these approaches failing to scale beyond datasets of the size of CIFAR-$10$. %

In contrast statistical methods typically leverage a process known as \emph{randomised smoothing}, in which repeated model 
draws under noise are employed to produce what is known as a \emph{smoothed classifier}, the properties of which can be exploited to construct guarantees of model robustness. This is made possible by attempting to parameterise worst-case behaviours under attack. While the addition of this noise is not cost-free, it is an embarrassingly parallel process that requires significantly fewer resources to scale to large models and complex datasets than are required with convex relaxation. Moreover, randomised smoothing does not require any modifications to the core model architecture, nor to the training and testing loops, which significantly reduces the level of engineering required to support the deployment of certified guarantees. It is due to these factors that for the remainder of this work we will only consider such statistical certification techniques. 

\subsection{Randomised Smoothing}\label{sec:randomised_smoothing}

To construct the smoothed classifier $g$, the model is exposed to repeated samples under noise $\mathcal{N}(\mathbf{x}, \sigma^2 \mathbf{I})$. However, rather than producing a stochastic model, by taking the expectation of the model outputs under noise, $g$ becomes deterministic a deterministic property, which can be translated into a certification by attempting to parameterise the worst-case response of the model to perturbations.

To date, a number of different parameterisation approaches have been proposed, producing certifications of varying tightness. However, these works often leverage different mechanisms to perform their smoothing, the nuance of which has not been fleshed out in prior works. To help formally distinguish between techniques, we will henceforth refer to techniques as drawing upon either the \emph{softmax} expectation (sometimes referred to as the soft expectation), which represents the expected model output under noise; or the \emph{multinomial} expectation (sometimes referred to as the hard expectation), which represents the expectation of the $\argmax$ of a models outputs, which is equivalent to the expected predicted class under noise. While conceptually similar, these two approaches can mathematically be respectively represented by way of $\tilde{\mathbf{Y}}$ and $\tilde{\mathbf{Y}}'$, where
\begin{align}
    \tilde{\mathbf{X}} & \sim \mathcal{N}(\mathbf{x},\sigma^2\mathbf{I}) 
   && & \tilde{\mathbf{Z}} & = f_{\boldsymbol{\theta}}(\tilde{\mathbf{X}}) \nonumber\\
    \tilde{\mathbf{Y}} & = \Pi(\tilde{\mathbf{Z}}) && & \tilde{Y}'_j & = \begin{cases}
1, &\tilde{Z}_j > \max_{k\in \mathcal{K}\setminus j}\tilde{Z}_k \\
0, &\mbox{otherwise\enspace,}
\end{cases}
\end{align}
were $\Pi(\cdot)$ represents the softmax operator. Deterministic expectations over these classes are estimated with high probability by constructing a Monte-Carlo estimate over $\mathcal{D}$, requiring $n$ i.i.d. draws either $(\tilde{\mathbf{x}}_i,\tilde{\mathbf{z}}_i,\tilde{\mathbf{y}}_i)$ or $(\tilde{\mathbf{x}}_i,\tilde{\mathbf{z}}_i,\tilde{\mathbf{y}}'_i)$. From this, the output of the smoothed classifier $g$ corresponds to the expectations over $\tilde{\mathbf{Y}}$ and $\tilde{\mathbf{Y}}'$, where %
\begin{eqnarray}
 \mathbb{E}_{\mathcal{D}}[\tilde{\mathbf{Y}}] = \frac{1}{n} \sum_{i=1}^{n} \tilde{\mathbf{y}}_i &&
\mathbb{E}_{\mathcal{D}}[\tilde{\mathbf{Y}}'] = \frac{1}{n} \sum_{i=1}^{n} \tilde{\mathbf{y}}'_i\enspace.
\end{eqnarray}
The stability of these expectations at inference time are supported by augmenting each training time sample with noise drawn from $\mathcal{N}(\mathbf{0}, \sigma^2 \mathbf{I})$.  %

Given that certification mechanisms seek to guarantee the behaviour of models under potential attack, the introduction of Monte-Carlo estimates of the expectation may appear to be inherently contradictory and unsuitable. However, if we are able to definitively calculate the worst-case expectations for a given Monte-Carlo samplings output, then any subsequent certification can still be confidently considered as a worst-case, conservative bound upon the existence of any potential adversarial examples. 

In the case of softmax expectations, the calculated and worst-case expectations are related by the well-known Hoeffding inequality \cite{hoeffding1994probability}, which provides a high-probability tail bound for a confidence level $\alpha > 0$, for which
\begin{equation}\label{eqn:SoftmaxBounds}
    P\left(\left|\mathbb{E}_{S}[\mathbf{Y}] - \mathbb{E}[\tilde{\mathbf{Y}}]\right| \leq \sqrt{\frac{\log_{e}(2/\alpha)}{2 n}}\right) \geq 1 - \alpha\enspace,
\end{equation}
for the Monte-Carlo estimate $\mathbb{E}[\tilde{\mathbf{Y}}]$ and true, worst case softmax expectation $\mathbb{E}_{S}[\mathbf{Y}]$. 

For multinomial output distributions, which we will henceforth label as $\mathbb{E}_{S}[\mathbf{Y}']$, we propose treating the two highest class outputs as distinct and unique outputs of a binomial distribution, and measuring uncertainties as such. Such bounds can be estimated by way of the Beta distribution, which reliably produces bounds that achieve the nominal coverage \cite{cameron2011estimation}. These uncertainties are calculated subject to the Bonferroni correction to $\alpha$ \cite{dunn1961multiple}, to account for the two measures being were drawn from the same sampling process. Taking this approach is significantly computationally cheaper than other, more comprehensive mechanisms for constructing bounds upon the expectations \cite{sison1995simultaneous, goodman1965simultaneous}, while still producing guaranteed coverage.

For future clarity, we will henceforth refer to the sorted softmax and multinomial class expectations as $E_S$ and $E_M$ respectively, where the first element of each---$E_{S,0}$ and $E_{M,0}$---employ the calculated lower bound on the estimated expectations, while the second element---$E_{S,1}$ and $E_{M,1}$---correspond to the calculated upper bounds.

\subsection{Certification Mechanisms}

While previous works have considered the softmax and multinomial expectations to be broadly interchangeable, it is important to emphasise that the conceptual differences between the value of these outputs means that they are addressing fundamentally similar-but-distinct problem spaces. As such, we will now summarise key certification mechanisms for $\ell_2$ threat models in a fashion that reflects the applicable expectation framework for the technique at hand.

The first randomised smoothing based certifications drew upon differential privacy~\cite{dwork2006calibrating} in order to bound the response of models under noise-based perturbation, leading to what is known as the Lecuyer \etal~\cite{lecuyer2019certified} approach. Under this framework, certified robustness over $\boldsymbol{\gamma} \in B_{p}(L)$ can be calculated by way of
\begin{align}\label{eqn:Lecuyer_bound}
&L_{\text{Lecuyer}} = \\
&\qquad \max_{\epsilon \in (0, 1]}\frac{\sigma \epsilon}{\bigtriangleup \sqrt{2 \log(1.25 (1 + e^{\epsilon} ) / (E_{S,0} - e^{2 \epsilon} E_{S,1}))}}\enspace.
\end{align}
Here $\bigtriangleup$ is a variant of the local Lipschitz continuity with respect to input perturbations of the base model $f(\cdot)$, which for $\ell_2$-norm-bounded perturbation corresponds to 
\begin{equation}
\bigtriangleup = \max_{\mathbf{x}, \mathbf{x} + \boldsymbol{\gamma}} \frac{\| f(\mathbf{x}) - f(\mathbf{x} + \boldsymbol{\gamma}) \|_2}{\| \mathbf{x} - (\mathbf{x} + \boldsymbol{\gamma} ) \|_2}\enspace.
\end{equation}
While Equation~\ref{eqn:Lecuyer_bound} does rely upon finding a maximum, failing to reach a global maxima does not void the certification, as the established bound is provably true for all $\epsilon$. 

In practice, while Lecuyer \etal explicitly framed this approach in terms of the softmax output distribution, it can be applied to systems which only return a multinomial output distribution. 

While the above certification mechanism was the first to provide guarantees of robustness for data sets as large as ImageNet, the conservative nature of the established bounds has left scope for new techniques to try and extend the size of achievable certifications. This was demonstrated by Li \etal~\cite{li2018certified}, who exploited R\'{e}nyi Divergence to provide an improved guarantee of size
\begin{equation}\label{eqn:Li_Bound}
L_{\text{Li}} = \sup_{\omega > 1} \sigma \sqrt{
\begin{aligned}
\log \left[\vphantom{\frac{1}{2}}\right. &1 - E_{M,0} - E_{M,1}
  + 2 \left( \frac{1}{2}E_{M,0}^{1-\omega}\right.\\
  &+ \left.\left.\frac{1}{2}E_{M,1}^{1 - \omega} \right)^{\frac{1}{1 - \omega}} \right]^{\frac{-2}{\omega}}
\end{aligned}  
  } \enspace.
\end{equation}
Unlike the approach of Lecuyer \etal, this this approach does not apply to outputs employing the softmax expectations.

The most popular mechanism in the current literature was developed by Cohen \etal~\cite{cohen2019certified, salman2019provably}, and constructs certifications in terms of the multinomial output by way of the Gaussian quantile function $\Phi^{-1}$, yielding certifications:%
\begin{equation}\label{eqn:Cohen_Bound}
L_{\text{Cohen}} = \sigma \left( \Phi^{-1}\left(E_{M,0}\right) \right)\enspace.
\end{equation}
While this work was presented alongside a second certification in terms of $E_{M,0}$ and $E_{M,1}$ that presents a tighter bound, their experiments exclusively considered the form above, which we will follow. It must also be noted that previous implementations of Equation~\ref{eqn:Cohen_Bound} have used a process based on the binomial distribution, which introduces a low probability chance of incorrectly selecting the wrong output class and producing a failed certification---a detail that is further discussed in Section~\ref{sec:implementation}. To alleviate these concerns we will consider \emph{Cohen} \etal to refer to Equation~\ref{eqn:Cohen_Bound} subject to the same multinomial distribution as Li \etal. %

We note that recent works have provided further extensions upon the radii of certification achievable through these mechanisms. Some attempt to improve the mechanisms through which we calculate certifications~\cite{cullen2022double}; while others attempt to induce shifts in the output distribution through training time loss-function modifications that incentivise larger certification radii~\cite{salman2019provably, zhang2019towards}, with MACER being particularly popular~\cite{zhai2020macer}. However, deploying all of these approaches introduce significant increases in the requisite computational time, with MACER inducing a $40$-fold increase in training time on our system. Moreover, it is crucial to emphasise that all of these systems for modifying training time certified robustness still derive their certifications using the approach of Cohen \etal. As such, within this work we will focus our improvements upon these core certification regimes, rather than their extensions, as improvements to the these core routines will still yield improvements when the modified mechanisms are deployed.

\section{Comparing Certification Performance}\label{sec:dataindpt}

Each of the aforementioned certification mechanisms has demonstrated its utility by showing improvements over the previously established state of the art. In each of these works, the core metric has been the certified accuracy, which corresponds to the proportion of samples that are correctly predicted with a radius greater than $r$, equivalent to
\begin{equation}
    c_A(r) = \frac{1}{N} \sum_{i=1}^{N} \mathds{1}[\argmax g(\mathbf{x}_i) = l(\mathbf{x}_i)] \mathds{1}[L(g(\mathbf{x}_i)) > r]\enspace,
\end{equation}
where $l(\mathbf{x})$ is the correct class label for the sample $\mathbf{x}$, and $L(\cdot)$ is the certification stemming from the smoothed classifier $g$. 

While such a measure allows for comparisons between techniques to be easily parsed, it also presents a picture that suggests that improvements in certification performance between approaches are uniform across all samples. In doing so, the drivers of certification performance in different techniques are distorted, and more difficult to interrogate. This is intrinsically problematic, as it both limits our ability to understand how a technique may generalise to new, semantically different datasets; and the capacity to assess certification performance for datasets where samples have differing adversarial risks.

In order to resolve these limitations in how we assess certifications schemes, we introduce a simple but powerful observation: \emph{certification mechanisms often decompose such that the only determinant of certification performance are the model output expectations}. This would appear to be immediately contradicted by Equations~\ref{eqn:Lecuyer_bound}, ~\ref{eqn:Li_Bound}, and ~\ref{eqn:Cohen_Bound}, however each of these exhibits linear, multiplicative proportionality to $\sigma$, and as such, the only determinant of performance is the expectation set. While this is an obvious statement, it is also not a consideration that any prior work has considered, and allows for a dataset, learner, and model agnostic mechanism for comparing certification approaches, while also providing a framework for disaggregated analysis of the performance of any specific combination of dataset, learner, and model. 

We introduce Definition~\ref{def:cert-compose} to formalise this statement, and to emphasise that certification mechanisms strictly depend upon the projection of the output $C_K$ space onto the $C_3$-simplex. Thus any comparison between technique performance can also be considered in terms of this permissible space, yielding what we term as a \emph{dataset independent comparison}. In doing so, the performance of the certifications can be compared in a fashion that is agnostic to the choice of model, dataset, or training infrastructure employed.

\begin{definition}\label{def:cert-compose}
Consider the set of smoothed classifiers $\mathcal{G}$, from which any smoothed classifier $g \in \mathcal{G}$ constructs the mapping $g: \reals^d \to C_K$ from input $\mathbf{x}$ to a point in the $K$-simplex, for $K$ output classes. A certification mechanism for the model family $\mathcal{G}$ is a mapping $h: \mathcal{G} \times \reals^d \to \reals^{\geq 0}$ from a model and input instance to a certified radius. 

\end{definition}

\begin{figure*}
    \begin{center}
        \includegraphics[width=0.35\textwidth]{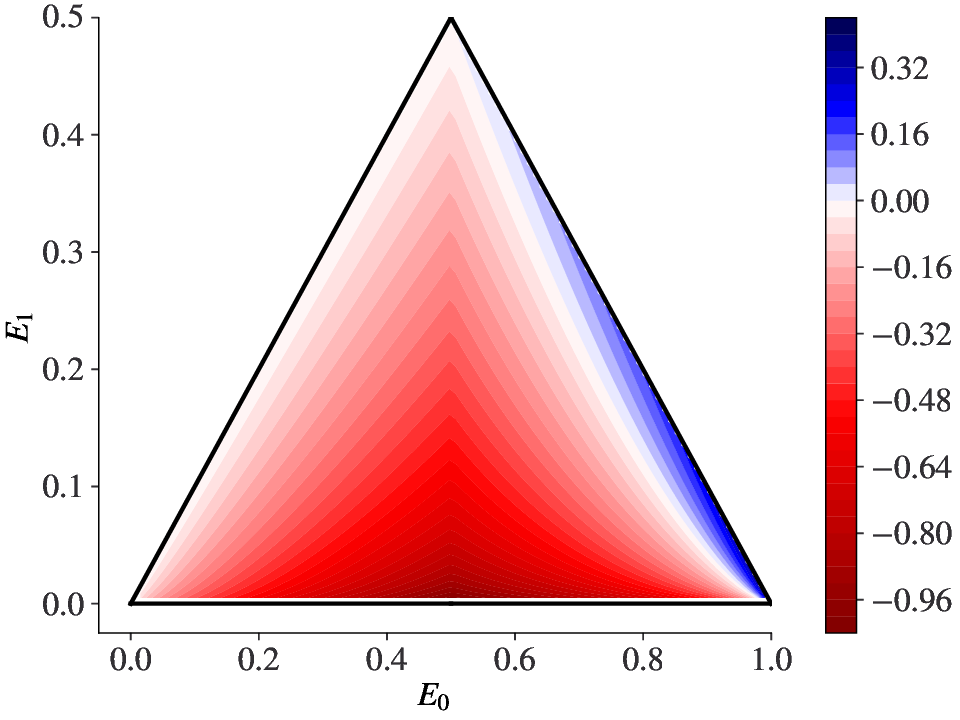}
    \end{center}
\caption{Comparing the relative performance of Equations~\ref{eqn:Li_Bound} and ~\ref{eqn:Cohen_Bound} by considering $(L_{\text{Cohen}}-L_{\text{Li}})$ at $\sigma=1.0$, by projecting all possible class outputs onto the surface of the $3$-simplex. Note that the linear multiplicative proportionality to $\sigma$ ensures that these relative scores can be rescaled by multiplying by some new $\sigma'$.}%
\label{fig:Cohen_vs_Li}    
\end{figure*}

To explore this concept, Figure~\ref{fig:Cohen_vs_Li} considers the relative certification performance between the Cohen \etal and Li \etal approaches, by considering the permissible space of expectations across the $C_3$ simplex as inputs, rather than the output of any specific model. While prior works have considered the Cohen \etal approach to uniformly produce the largest achievable radii of certification, in practice the commonly employed form of Cohen only out-certifies in the neighbourhood of $E_0 + E_1 \approx 1$. Outside of this region---which is likely to be seen in datasets which exhibit significant semantic overlap between classes---Li \etal begins to produce significantly larger certifications.

\subsection{Distributionally Aware Comparisons}

The above framing demonstrates that the performance of certification schemes can be considered strictly in context of the permissible output space in $C_3$. Doing so allows for a direct comparison of the performance of certification schemes without relying upon the model, dataset, or any other parts of the learning infrastructure. However, Definition~\ref{def:cert-compose} also suggests a second feasible framework for assessing certification performance. Consider the output distribution of $g(\mathbf{x})$ where the samples $\mathbf{x}$ are drawn from some data distribution $\mathcal{P}$. Assessing this output distribution in the context of Figure~\ref{fig:Cohen_vs_Li}, the dataset specific drivers of certification performance can be considered. It is this form of comparison that we will refer to as a \emph{distributionally-aware} analysis of certification performance. 

Such a perspective is valuable as it inherently allows us to better understand the factors driving certification performance for a particular trained model. In doing so also allows for inferences to be made about the potential certification performance of new and untested datasets, based upon their semantic complexity. Such an analysis may also allow deployed systems to develop an understanding of the risk of attack for particular samples.

\section{An Improved Differential Privacy Based Mechanism}\label{sec:improved_mechanism}

Extending the dataset-independent analysis of Figure~\ref{fig:Cohen_vs_Li} to incorporate Lecuyer \etal reveals that for multinomial outputs it is uniformly outperformed across the entirety of permissible outputs. However, the very notion that differential performance is possible upon a sample-wise basis suggests that improving upon the bounds delivered by Lecuyer \etal may achieve improvements over a subset of the output space. This is especially true as recent works have shown the differential privacy mechanism that underpins Lecuyer \etal underestimates the level of privacy that can be achieved for a given level of added noise~\cite{balle2018improving, zhao2019reviewing}. Within the remainder of this section, we will demonstrate how this improved mechanism can be incorporated into a new certification regime, that both uniformly outperforms Lecuyer \etal across the $C_3$ simplex; and yields improvements over Li \etal and Cohen \etal for the majority of the permissible output space.

In aide of this goal, we begin by introducing some core concepts of differential privacy: differential privacy as a stability condition on output distributions and how it translates to the stability of expected outputs (\Cref{lem:exp-stability}); the post-processing inequality (\Cref{lem:post-process}) and how it captures the invariance of differential privacy to data-independent compositions; and the improved analysis of the Gaussian mechanism. Of these, \Cref{lem:exp-stability} and \Cref{lem:post-process} follow Lecuyer \cite{lecuyer2019certified}; and the improved analysis of the Gaussian mechanism follows \cite{balle2018improving}.

\begin{lemma}[\textbf{Expected Output Stability Bound}]\label{lem:exp-stability}
Consider a randomised function $A:
\reals^n \to [0,1]$ that
preserves $(\epsilon,\delta)$-DP, then it must be that $\mathbb{E}[A(\mathbf{x})]\leq e^\epsilon \mathbb{E}[A(\mathbf{x}+\boldsymbol{\gamma})] + \delta,$ 
where the expectations are taken over the randomness in $A$.
\end{lemma}

The familiar post-processing inequality of differential privacy~\cite{dwork2006calibrating} is critical for certification in that it permits privacy-preserving randomisation to be applied at early network layers.

\begin{lemma}[\textbf{Post-Processing Inequality}]\label{lem:post-process}
Consider any randomised algorithm $A$ acting on databases, and any (possibly randomised) algorithm $B$ with domain $range(A)$. If $A$ is $(\epsilon,\delta)$-DP then so too is $B\circ A$. Moreover, at the level of database pairs $R_{\epsilon,\delta}(A)\subseteq R_{\epsilon,\delta}(B\circ A)$.
\end{lemma}

The $\epsilon$-DP of a random mechanism $A(\mathbf{x})$ is captured by the privacy loss random variable
\begin{equation}l_{A,\mathbf{x}, \mathbf{x}'} = \log \left( \frac{P(A(\mathbf{x}) = \mathcal{O})}{P(A(\mathbf{x}') = \mathcal{O})} \right)\enspace,\end{equation}
where $\mathbf{x}' = \mathbf{x} + \boldsymbol{\gamma}$. By then introducing $L_{A, \mathbf{x}, \mathbf{x}'} = l_{A, \mathbf{x}, \mathbf{x}'}(Y)$ \cite{balle2018improving}, an equivalent condition for differential privacy is
\begin{equation}\label{eqn:PRLV_DP}
P[L_{A, \mathbf{x}, \mathbf{x}'} \geq \epsilon] - e^{\epsilon} P[L_{A, \mathbf{x}, \mathbf{x}'} \leq -\epsilon] \leq \delta\enspace.
\end{equation}
To elaborate upon these probabilities, consider a mechanism of the form $y \sim h(\mathbf{x}) + \mathcal{N}(0, \sigma^2 \mathbf{I})$, where $h(x)$ is any arbitrary function. Taking such a framing allows the privacy loss random variable to be analytically expressed as
\begin{align}
l_{A,\mathbf{x}, \mathbf{x}'} &= -\frac{1}{2 \sigma^2}\left(\norm{\mathbf{y} - h(\mathbf{x})}_2^2 - \norm{\mathbf{y} - h(\mathbf{x}')}_2^2 \right)\enspace,\nonumber\\
&= \frac{1}{2 \sigma^2}\norm{h(\mathbf{x}) - h(\mathbf{x}')}_2^2 + \frac{1}{\sigma^2}\langle \mathbf{y} - h(\mathbf{x}), \\
&\hspace{3.5 cm} h(\mathbf{x}) - h(\mathbf{x}') \rangle\enspace\nonumber.
\end{align}
A consequence of the fact that the inner product is equivalent to $\mathcal{N}\left(0, \sigma^2 \norm{h(\mathbf{x}) - h(\mathbf{x}')}_2^2\right)$ is that
\begin{equation}A_{\mathbf{x}, \mathbf{x}'} = \mathcal{N}\left(\frac{\norm{h(\mathbf{x}) - h(\mathbf{x}')}_2^2}{2 \sigma^2}, \frac{\norm{h(\mathbf{x}) - h(\mathbf{x}')}_2^2}{\sigma^2}\right)\enspace.\end{equation}
Based upon this framing, the components of Equation~\eqref{eqn:PRLV_DP} can be constructed as
\begin{align}\label{eqn:bounds_sigma}
&P[L_{A, \mathbf{x}, \mathbf{x}'} \geq \epsilon] = \nonumber \\
&\qquad \Phi \left(\frac{\norm{h(\mathbf{x}) - h(\mathbf{x}')}_2}{2 \sigma} - \frac{\epsilon \sigma}{\norm{h(\mathbf{x}) - h(\mathbf{x}')}_2}\right)\enspace,\nonumber\\
&P[L_{A, \mathbf{x}, \mathbf{x}'} \leq -\epsilon] = \\
&\qquad \Phi \left(-\frac{\norm{h(\mathbf{x}) - h(\mathbf{x}')}_2}{2 \sigma} - \frac{\epsilon \sigma}{\norm{h(\mathbf{x}) - h(\mathbf{x}')}_2}\right)\enspace, \nonumber
\end{align}

Extending this concept to certified robustness requires the application of the post-processing inequality of~\ref{lem:post-process}. If we consider a function $h(\cdot)$ such that $h(\mathbf{x}) = \mathbf{x}$, then Equations~\ref{eqn:PRLV_DP} and \ref{eqn:bounds_sigma} can be combined to take the form

\begin{equation}\label{eqn:TightDPNoise}
    \Phi\left(\frac{L}{2 \sigma} - \frac{\epsilon \sigma}{L} \right) - e^{\epsilon} \Phi\left(-\frac{L}{2 \sigma} - \frac{\epsilon \sigma}{L} \right) \leq \delta\enspace, 
\end{equation}

where $L = \| \mathbf{x} - \mathbf{x}'\|$ is the certified radius. If this is true for any function $y \sim h(\mathbf{x}) + \mathcal{N}(0, \sigma^2 \mathbf{I})$, then by virtue of the post-processing inequality the equivalent privacy relationship also holds for any mechanism $f(\cdot)$, which allows us to define our randomised mechanism as
\begin{align}
A(\mathbf{x}) &= f(h(\mathbf{x}) + \mathcal{N}(0, \sigma^2 \mathbf{I})) \nonumber \\
&= f(\mathbf{x} + \mathcal{N}(0, \sigma^2 \mathbf{I}))\enspace.
\end{align}
This definition of $A(\mathbf{x})$ then becomes equivalent to the $g(\mathbf{x})$ of Section~\ref{sec:randomised_smoothing}, if the expectations were to be taken over a single draw of noise. 

In a similar fashion to Equation~\ref{eqn:Lecuyer_bound}, this differential privacy based certification scheme can be framed as a maximisation problem, especially as Equation~\ref{eqn:TightDPNoise} does not admit an analytic inverse. However, rather than strictly considering $\epsilon \in (0,1]$ as the optimisation criteria, the above criteria can be recast as a constrained optimisation problem over $\epsilon \geq 0$ and $\delta \in [0,1]$ in order to construct a certification by way of

\begin{align}\label{eqn:NewDP}
    \begin{split}
        L &= \max_{\epsilon \geq 0, \delta \in [0,1]} L'\\
        &\textrm{s.t. } \Phi\left(\frac{\bigtriangleup L'}{2 \sigma} - \frac{\epsilon \sigma}{\bigtriangleup L'} \right) - e^{\epsilon} \Phi\left(-\frac{\bigtriangleup L'}{2 \sigma} - \frac{\epsilon \sigma}{\bigtriangleup L'} \right) \leq \delta\enspace, \\
&E[A_{i}(\mathbf{x})] - \max_{j \in \mathcal{K} \setminus \{i\}} E[A_{j}(\mathbf{x})] e^{2\epsilon} - 
          (1 + e^{\epsilon}) \delta \geq 0
          \enspace. %
    \end{split}
\end{align} %
where $i$ corresponds to the predicted class, and $E[A_{i}(\mathbf{x})]$ and $E[A_{j}(\mathbf{x})]$ are respectively upper and lower bounded as per Section~\ref{sec:randomised_smoothing}. While the form of the above equation is complex and nonlinear, the constraint functions exhibit near-monotonic behaviour in $(\epsilon, \delta, L)$, and as such can be quickly solved with any conventional constrained numerical optimisation tools. That this approach is only conditional upon $A(\mathbf{x}) \in [0,1]$, it can be applied to both softmax and multinomial distributions, or indeed any $A(\mathbf{x}) \in [0,c]$, as the latter case is simply a uniform scaling of $[0,1]$. The provably true nature of these guarantees is a direct consequence of Lecuyer \etal~\cite{lecuyer2019certified}, as our approach tightens their bounds. 

Beyond its ability to incorporate the improved privacy mechanism, framing the certification process as an optimisation process presents an additional advantage over the differentially privacy certifications of Lecuyer \etal: we remove the need to arbitrarily set the $(\epsilon, \delta)$-privacy level prior to certification. This is meaningful as most of the contexts for which certification is useful do not care for a specific, fixed privacy level across all samples, rather they value producing the largest achievable certification, in order to more accurately gauge adversarial risks.

\begin{figure*} 
    \centering
  \subfloat[Softmax]{%
       \includegraphics[width=0.38\linewidth]{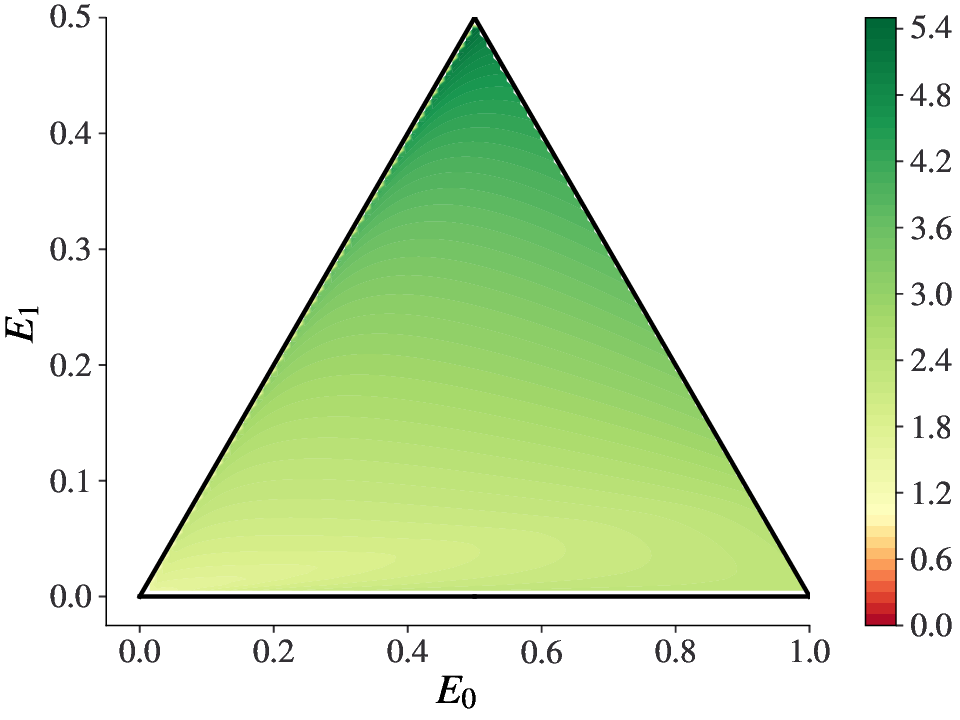}}
    \hfill
  \subfloat[Multinomial]{%
        \includegraphics[width=0.38\linewidth]{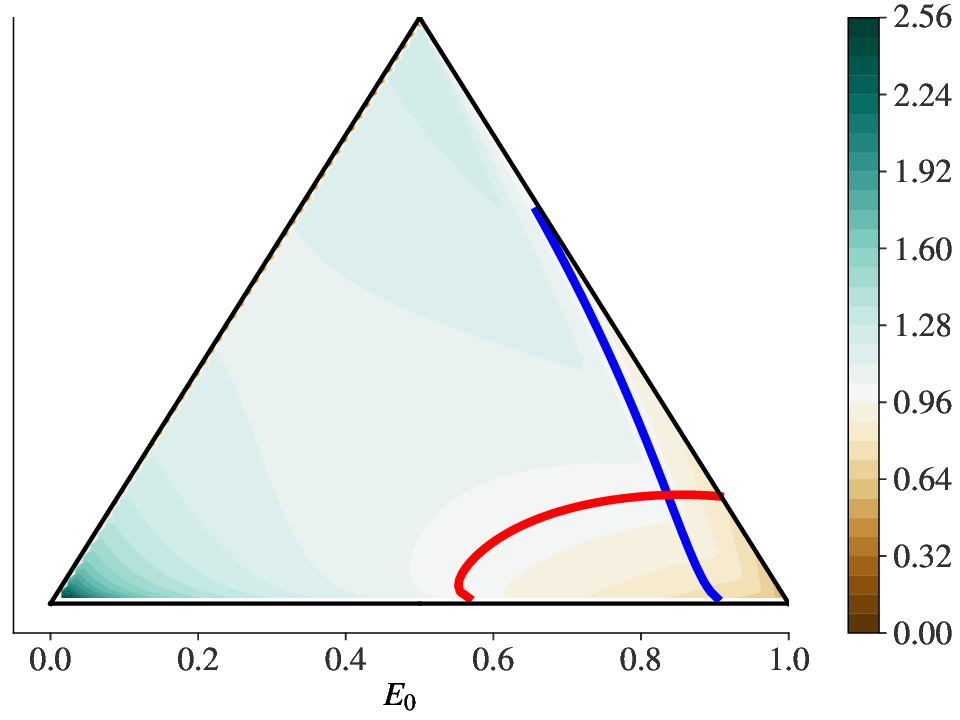}}
\caption{Assessing the relative performance of our technique by considering the ratio between $L_{\text{Ours, Softmax}}$ and $L_{\text{Lecuyer, Softmax}}$ for the Softmax distribution; while the Multinomial figure considers the ratio between $L_{\text{Ours, Multinomial}}$ and the maximum of $L_{\text{Lecuyer, Multinomial}}, L_{\text{Li}},$ and $L_{\text{Cohen}}$. In the Multinomial figure, samples to the right of the blue and red lines respectively refer to when Cohen \etal and Li \etal produce the largest certification, following Figure~\ref{fig:Cohen_vs_Li} . }\label{fig:Theoretical_comparison_results}
\end{figure*}

In the context of the dataset-agnostic comparisons across the $C_3$ simplex, Figure~\ref{fig:Theoretical_comparison_results} reveals that for a softmax output distribution, our approach uniformly outperforms Lecuyer \etal across the entire simplex, exhibiting a maximal relative $5$-fold improvement in the calculated certification.

For a multinomial output distribution, our new technique yields improved certifications against both Cohen \etal and Li \etal over the majority of the output space. When incorporating our technique into the comparison, across the $C_3$ simplex Cohen \etal produces strong certifications only in the neighbourhood of  $E_1 = 1 - E_0$; while  Li \etal~\cite{li2018certified} produces the strongest certification near $(E_0, E_1 )  \to (1, 0)$. However, as $E_0$ and $E_1$ both decrease---as seen in semantically complex samples---our comparisons demonstrate that it is possible to inrease the certification more than two-fold. %

\subsection{Cost-Free Improvements To Certifications}\label{sec:ensemble_mechanisms}

Our next key observation builds upon Section~\ref{sec:dataindpt}: \emph{if each base certification mechanism is superior to all other mechanisms under consideration even on only one point in the output simplex, then taking the maximum across an ensemble of mechanisms' radii provably dominates the performance of any single mechanisms}.

\begin{corollary}[\textbf{Ensembling Certifications}]\label{cor:ensemble}
Consider the set of certification mechanisms $L_1,\ldots,L_n$, each of which incorporate a mapping from the $L_i : C_3 \to \reals^{\geq 0}$. Each of these yields a certification $k_i(g, \cdot) = (L_i \cdot g)(\cdot)$, where $k_i : \mathcal{R}^d \to \reals^{\geq 0}$. If $L'(\mathbf{s}) = \max_{i \in {1, \ldots, n}} L_i(\mathbf{s})$ then $k(g, \mathbf{x}) \geq k_i(g, \mathbf{x})$ for all $g \in \mathcal{G}, \mathbf{x} \in \reals^d$. Moreover, if each region of superiority each region of superiority $H_i=\{\mathbf{s}\in C : k_i(\mathbf{s}) > k_j(\mathbf{s}), \forall j\neq i\}$ is non-empty, then $k$ dominates each base mechanism $g_1,\ldots,g_n$.

\end{corollary}

Figure~\ref{fig:CVPR} diagrammatically represents this ensembling mechanism, while the differential performance of Figure~\ref{fig:Theoretical_comparison_results} demonstrates both the nature of the elements of $H_i$ and the functional differences that empower the ensembling process. It must be stressed is that if certifications in terms of a softmax output distribution are sought then only our technique and that of Lecuyer \etal can be certified. However, if we certify in terms of the multinomial distribution, all of the randomised smoothing based techniques can be applied, although in practice the Lecuyer approach is uniformly outperformed by all other multinomial approaches.

It is important to emphasise that this ensemble certification process is almost cost-free, as the dominant computation in certification is evaluating $f(\mathbf{x})$ by Monte-Carlo estimation, and as such the incremental cost of ensembling by Corollary~\ref{cor:ensemble} is minimal, as all the techniques build upon the same expectations. The evaluation of subsequent $h_i(\cdot)$ involves a handful of arithmetic calculations or simple numerical library calls (for Normal quantiles), which is trivial by comparison to the cost of completely restarting the certification process from scratch.

Reusing the expectations across the ensembling process also obviates the need to adjust the confidence intervals, even though multiple experiments are being performed. This stems from the fact that we are calculating expectation ranges with high probability, and the worst-case variant of these is being applied to the certification mechanisms. That these certification mechanisms are deterministic interpretations of said expectations eliminates any considerations regarding potential multiple hypothesis testing. 

The concept of ensembling can be further extended by incorporating the convex relaxation style certifications as described in Section~\ref{sec:related}. However, while the randomised smoothing mechanisms share the majority of their computational burden between the techniques, no such opportunities to optimise the aggregate performance exist for convex relaxation methods. Due to this consideration, and the broader limitations of the convex relaxation based techniques, within this work we restrict our focus to those approaches that leverage randomised smoothing.

While contemporaneous work~\cite{yang2021certified} has considered  the robustness of ensembling neural network \emph{models}, our work considers ensembles of \emph{certification mechanisms} acting on a single common model. An ensemble of models requires multiple independent (and often costly) training loops, before requiring independent evaluations for each constituent model. In contrast, an ensemble of certification mechanisms allows for the majority of the computational burden of certification to be recycled between techniques, with the only additional burden the computational cost associated with solving the analytic certification equations.

\section{Implementation}\label{sec:implementation}

\begin{algorithm}
  \caption{Implementation of all tested algorithms under the Multinomial framework. Here $O-M$ denotes Our Multinomial approach.}\label{alg:Multinomial-C}
  \begin{algorithmic}[1]
    \Function{Multinomial-Certify}{$n, \mathcal{K}, \sigma, \alpha, \mathbf{x}, f$} \Comment{$n$ samples for class selection and certification; $\mathcal{K}$ set of output classes, $\sigma$ s.d of noise; $\alpha$ percentage confidence; $\mathbf{x}, f$ sample and function}
    \State $E = \frac{\text{Count}(n, \mathcal{K}, \sigma, \mathbf{x}, f)}{n}$
    \State $j = \argmax E$
    \State $E = \text{Sort} (E)$ \Comment{Sort in descending order}
    \State $E_0 = E[0] - \text{Confint}(E[0] \times n, n, \alpha)$ \Comment{Largest element of $E$. \textit{Confint} using Beta-method}
    \State $E_1 = E[1] + \text{Confint}(E[1]\times n, n, \alpha)$
    \State \textbf{if} $E_0 > 0.5$ \textbf{then} $L_{\text{C}} = $ Equation~\ref{eqn:Cohen_Bound} \textbf{else} 0
    \If{$E_0 > E_1$}
    \State $L_{\text{Li}} =$ Equation~\ref{eqn:Li_Bound} %
    \State $L_{\text{O-M}} = $  Equation~\ref{eqn:NewDP} %
    \Else
    \State $L_{\text{Li}} = 0, \qquad L_{\text{O-M}} = 0$
    \EndIf
    \State $L_{\text{Ensemble}} = \max \left(L_{\text{C}}, L_{\text{Li}}, L_{\text{O-M}} \right)$
    \State \Return $j, \left(L_{\text{Ensemble}}, L_{\text{C}}, L_{\text{Li}}, L_{\text{O-M}} \right)$
    \EndFunction
    \\
    \Function{Count}{$n, \mathcal{K}, \sigma, \mathbf{x}, f$}
    \State $c_j = 0 \text{ } \forall j \in \mathcal{K}$
    \ForAll{$i \in n$}
    \State $c_j = c_j + 1$ \textbf{if} $\argmax f(\mathbf{x} + \mathcal{N}(\boldsymbol{0}, \sigma^2 \mathbf{I})) = j$
    \EndFor
    \State \Return $c_j$
    \EndFunction
  \end{algorithmic}
\end{algorithm}
\begin{algorithm}[tb]
  \caption{Softmax Certification framework. Here $\Pi(\cdot)$ represents the softmax operator and $O-S$ denotes Our-Softmax implementation.}\label{alg:Softmax-C}
  \begin{algorithmic}[1]
    \Function{Softmax-Certify}{$n, \sigma, \alpha, \mathbf{x}, f$}\Comment{$n$ samples for class selection and certification; $\mathcal{S}$ set of output classes, $\sigma$ s.d of noise; $\alpha$ percentage confidence; $\mathbf{x}, f$ sample and function}
    \State $E =\frac{1}{n} \sum_{i=1}^{n} \Pi(f(\mathbf{x} + \mathcal{N}(\boldsymbol{0}, \sigma^2 I))$ %
    \State $j = \argmax E$
    \State $E = \text{Sort} (E)$
    \State $E_0 = E[0] - \text{Hoeffding}(E[0] \times n, n, \alpha)$
    \State $E_1 = E[1] + \text{Hoeffding}(E[1] \times n, n, \alpha)$
    \If{$E_0 > E_1$}
    \State $L_{\text{Lec}} = $ Equation~\ref{eqn:Lecuyer_bound} %
    \State $L_{\text{O-S}} = $ Equation~\ref{eqn:NewDP} %
    \Else
    \State $L_{\text{Lec}} = 0, \qquad L_{\text{O-S}} = 0$
    \EndIf
    \State \Return $j, \left(L_{\text{Lec}}, L_{\text{O-S}}\right)$
    \EndFunction
  \end{algorithmic}
\end{algorithm}

To demonstrate how the aforementioned processes can be implemented, Algorithms~\ref{alg:Multinomial-C} and \ref{alg:Softmax-C} cover certifications across all of the techniques for all of the multinomial and softmax approaches. For a given test sample $x$, the functions \textit{Multinomial-Certify} and \textit{Softmax-Certify} return the predicted class and certifications, with both tasks performed through randomised smoothing. Within this algorithm, the function \textit{Confint} refers to the Beta function approach with Bonferroni correction, as is described within Section~\ref{sec:randomised_smoothing}. %

\begin{algorithm}
  \caption{Cohen Original}\label{alg:Cohen_Original}
  \begin{algorithmic}[1]
    \Function{Binomial-Certify}{$n_0, n, \mathcal{K}, \sigma, \alpha, \mathbf{x}, f$} \Comment{$n_0$ and $n$ ($n_0 \ll n$) are samples for class selection and certification; $\mathcal{S}$ set of output classes, $\sigma$ s.d of noise; $\alpha$ percentage confidence; $\mathbf{x}, f$ sample and function}
    \State $j = \argmax \text{Count}(n_0, \mathcal{K}, \sigma, \mathbf{x}, f)$
    \State $E_0 = \frac{\text{Count}(n, j, \sigma, \mathbf{x}, f)}{n}$ \Comment{Returns a single value}
    \State $E_0 = E_0 - \text{Confint}(E_0 \times n, n, \alpha)$
    \If{$E_0 > 0.5$}
    \State $L_{\text{C}} = \sigma \Phi^{-1}(E_0)$
    \Else
    \State $L_C = 0$
    \EndIf
    \Return $j, L_C$
    \EndFunction
  \end{algorithmic}
\end{algorithm}

Every effort was made to accurately recreate the implementations of Li \etal and Lecuyer \etal, in order to perform a fair comparison. However, as was alluded to in Section~\ref{sec:related}, the approach of Cohen \etal has the potential to incorrectly classify samples, and in doing so fail to certify. This stems from \cref{alg:Cohen_Original}'s sampling approach, in which classification is based on a small number of samples---as few as $n_0 = 100$ in Cohen \etal, before a second sampling draw over $n$ samples estimates the expectation based solely upon the count of times the classifier's class is selected. 

While selecting a large enough $n$ produces tight bounds on the uncertainties for the expectation, the uncertainties for the $100$ samples are often large enough to not be able to accurately classify the output solution. Moreover, the very nature of the binomial sampling of Cohen---in which the expectation of the classifier's output is compared to the likelihood of the intersection of \textit{all other classes}---means that if the incorrect class is chosen, the mistake cannot be rectified without completely re-sampling. As such, we instead selected the expectation for Cohen based not upon the binomial sampling process, but the same multinomial sampling process employed by both us and Li \etal.

\begin{figure*} 
    \centering
  \subfloat[$\sigma=1.0$]{%
       \includegraphics[width=0.4\linewidth]{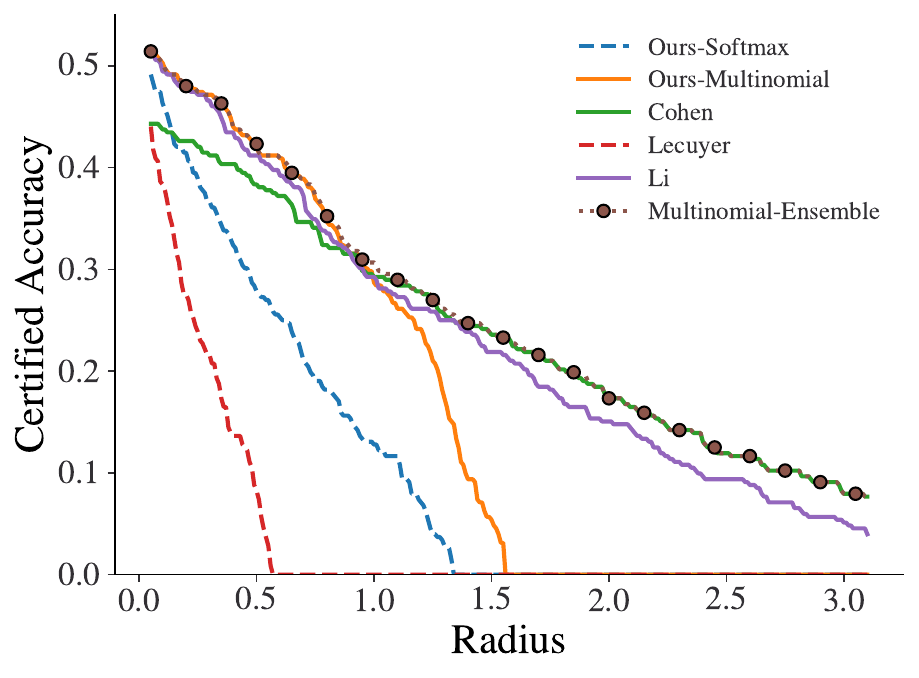}}
    \hfill
  \subfloat[Varying $\sigma$]{%
        \includegraphics[width=0.4\linewidth]{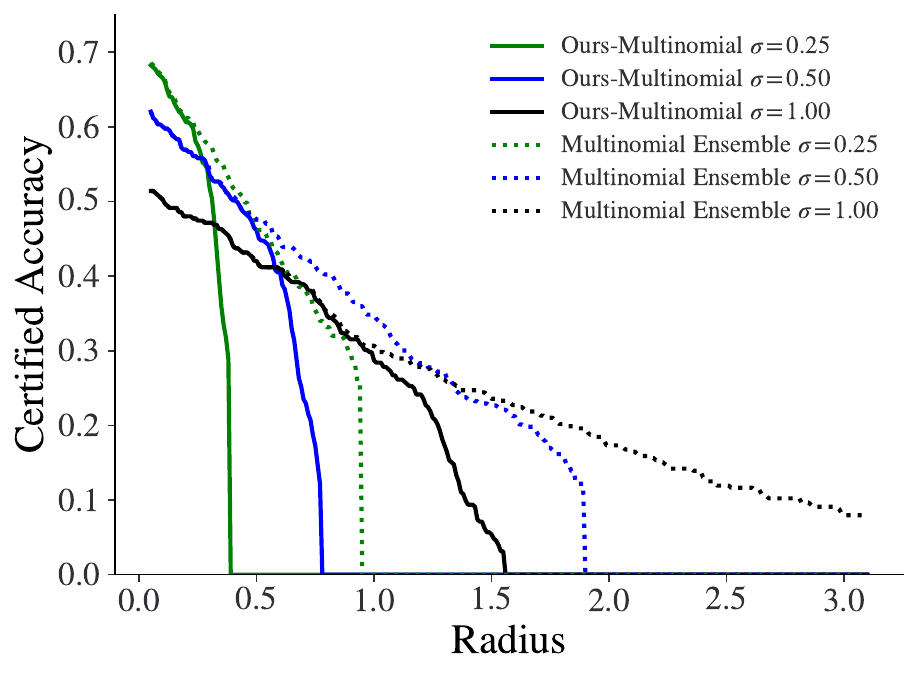}}
\caption{ImageNet Certified Accuracy. Figure a) demonstrates the variable performance of each technique, whereas the right considers the performance of our technique against the Ensemble (that includes our approach).}
  \label{fig:Imagenet}
\end{figure*}

\begin{figure*} 
    \centering
  \subfloat[Softmax]{%
       \includegraphics[width=0.42\linewidth]{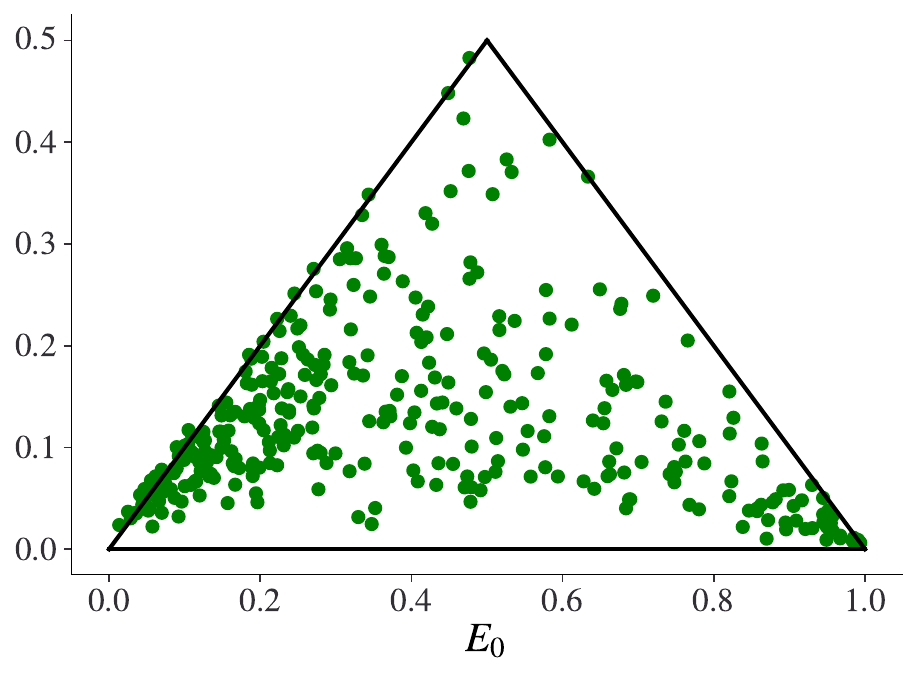}}
    \hfill
  \subfloat[Multinomial]{%
        \includegraphics[width=0.42\linewidth]{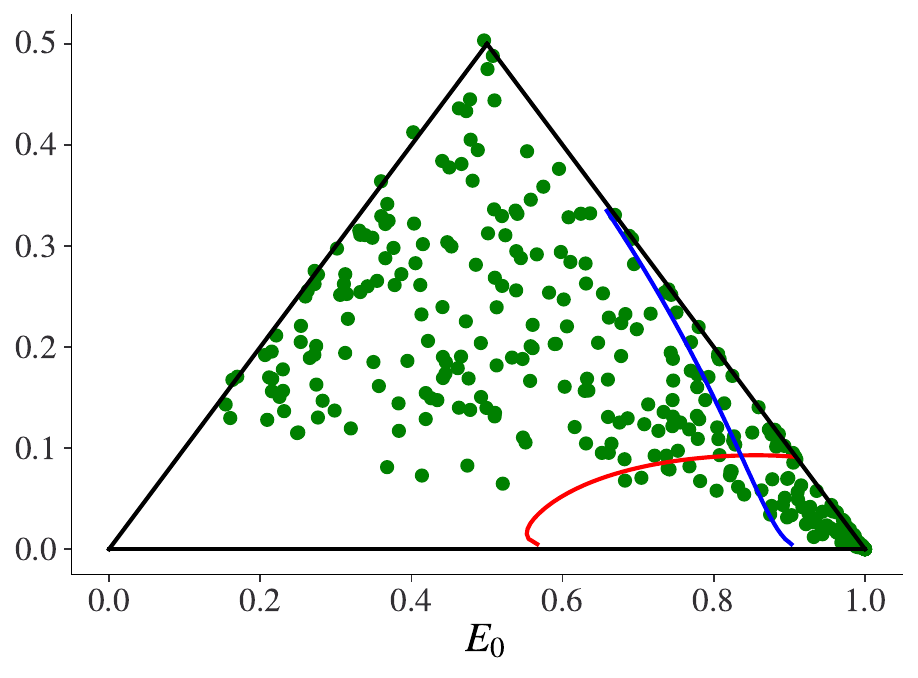}}
\caption{Distribution of output expectations when the samples are drawn from ImageNet, subject to the parameters $\sigma = 1$ and  $n = 100,000$. For Multinomial outputs samples to the right of the \emph{red} line are located in the region where Li \etal produces the largest certifications, while samples to the right of the \emph{blue} line are contained within the region where Cohen \etal outperforms all other techniques, as per Figure~\ref{fig:Theoretical_comparison_results}. Points outside these regions have their certifications improved upon when our technique is employed, including the entirety of the Sotmax output.} %
\label{fig:Scatters}
\end{figure*}

\begin{figure*} 
    \centering
  \subfloat[Certified Proportion \label{fig:IMAGENET_Proportion}]{%
       \includegraphics[width=0.46\linewidth]{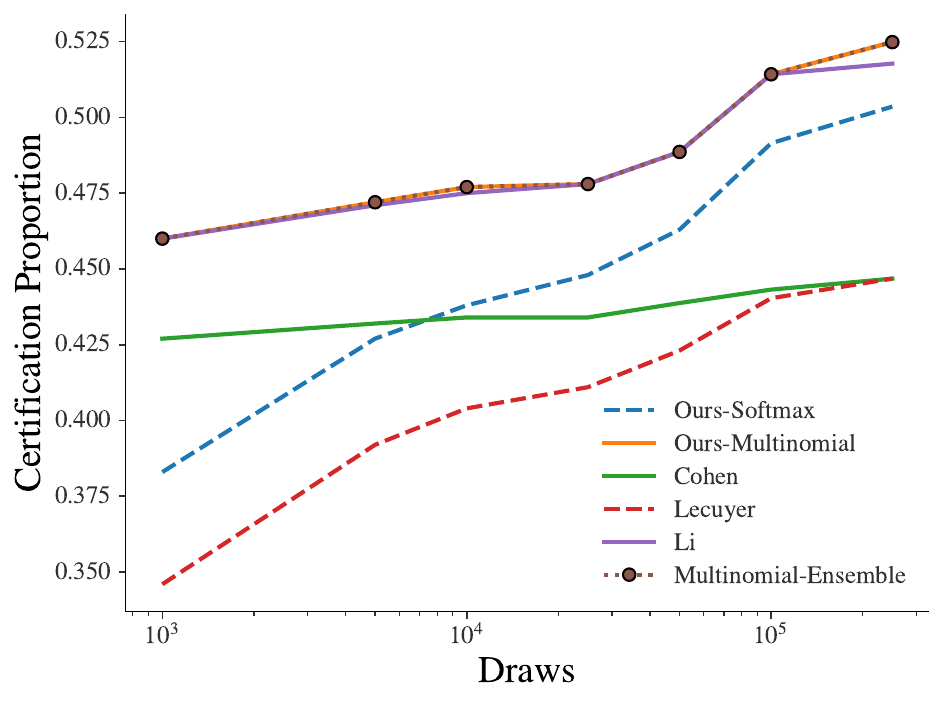}}
    \hfill
  \subfloat[Certification Time \label{fig:IMAGENET_Time}]{%
        \includegraphics[width=0.46\linewidth]{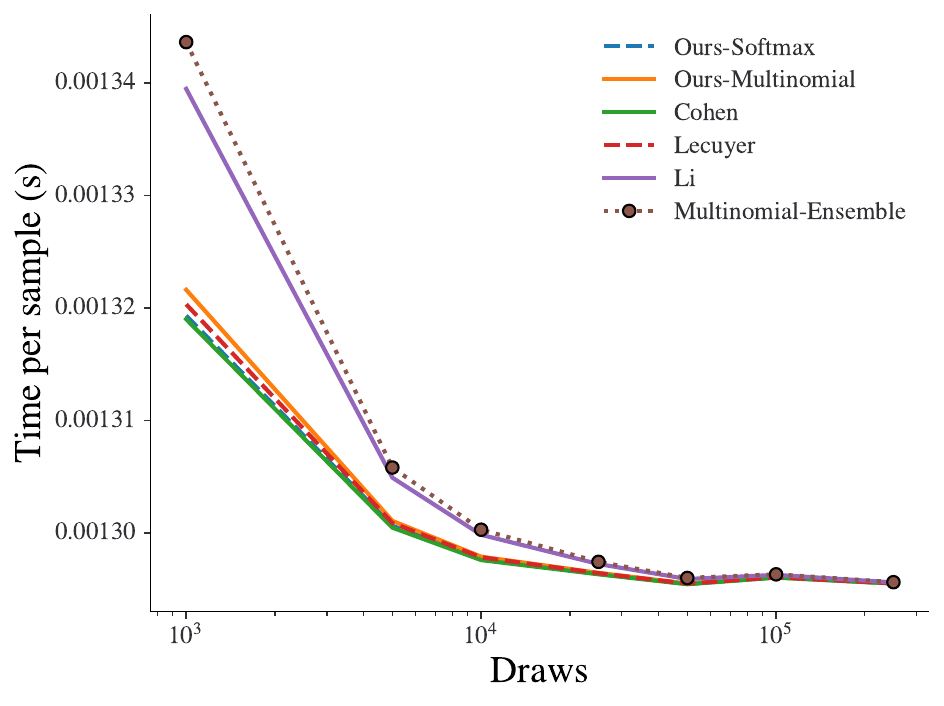}}
\caption{Certification Proportion (defined as $r > 0.05$) and Time for ImageNet at $\sigma = 1.0$. For Figure b), all times calculated were averaged over $100$ samples.}
  \label{fig:Imagenet_scaling}
\end{figure*}

\begin{table*}
\caption{ImageNet performance for $n = 10,000$ across a range of $\sigma$. Metrics cover the Top$-1$ accuracy (T1), median certification (M), the proportion of samples for which a technique produces the largest certification (L), and the proportion of samples for which $r > 0.05$, as an additional measure of significance. The strongest measure for each without-ensembling has been bolded, as well as all metrics in the Ensemble columns as by its nature it must  outperform any individual technique.}%
\label{tab:large_table}
  \centering
  \begin{tabular}{lllllllllllll}
    \toprule
 $\sigma$ & & \multicolumn{3}{c}{\textbf{Cohen}} & \multicolumn{3}{c}{\textbf{Li}} & \multicolumn{3}{c}{\textbf{Ours}} & \multicolumn{2}{c}{\textbf{Ensemble}} \\
 & T$1$ & M. & L. & $>$ & M. & L. & $>$ & M. & L. & $>$ & M. & $>$ \\
\cmidrule(r){1-1} \cmidrule(r){2-2} \cmidrule(r){3-5} \cmidrule(r){6-8} \cmidrule(r){9-11} \cmidrule(r){12-13} 
$\mathbf{1/4}$ & $0.64$ & $\mathbf{0.77}$ & $\mathbf{0.89}$ & $0.94$ & $0.61$ & $0.01$ & $\mathbf{0.96}$ & $0.35$ & $0.11$ & $\mathbf{0.96}$ & $\mathbf{0.77}$ & $\mathbf{0.96}$ \\
$\mathbf{0.5}$ & $0.49$ & $\mathbf{1.18}$ & $\mathbf{0.80}$ & $0.93$ & $1.02$ & $0.03$ & $\mathbf{0.97}$ & $0.67$ & $0.18$ & $\mathbf{0.97}$ & $\mathbf{1.18}$ & $\mathbf{0.97}$ \\
$\mathbf{1}$ & $0.43$ & $\mathbf{1.57}$ & $\mathbf{0.66}$ & $0.89$ & $1.40$ & $0.06$ & $0.97$ & $1.20$ & $0.28$ & $\mathbf{0.98}$ & $\mathbf{1.59}$ & $\mathbf{0.98}$ \\ 
$\mathbf{2}$ & $0.27$ & $1.49$ & $0.37$ & $0.76$ & $1.69$ & $0.17$ & $\mathbf{0.97}$ & $\mathbf{1.75}$ & $\mathbf{0.49}$ & $\mathbf{0.97}$ & $\mathbf{1.76}$ & $\mathbf{0.97}$ \\
$\mathbf{3}$ & $0.17$ & $0.82$ & $0.23$ & $0.62$ & $1.68$ & $0.22$ & $\mathbf{0.94}$ & $\mathbf{1.81}$ & $\mathbf{0.65}$ & $\mathbf{0.94}$ & $\mathbf{1.81}$ & $\mathbf{0.94}$ \\
$\mathbf{4}$ & $0.12$ & $0.06$ & $0.17$ & $0.5$ & $1.90$ & $0.22$ & $\mathbf{0.93}$ & $\mathbf{2.06}$ & $\mathbf{0.72}$ & $\mathbf{0.93}$ & $\mathbf{2.06}$ & $\mathbf{0.93}$ \\
$\mathbf{5}$ & $0.09$ & $0.00$ & $0.17$ & $0.42$ & $1.87$ & $0.17$ & $\mathbf{0.91}$ & $\mathbf{2.07}$ & $\mathbf{0.79}$ & $\mathbf{0.91}$ & $\mathbf{2.07}$ & $\mathbf{0.91}$ \\
    \bottomrule
  \end{tabular}
\end{table*}

\begin{figure}\begin{center}
    \includegraphics[width=0.38\textwidth]{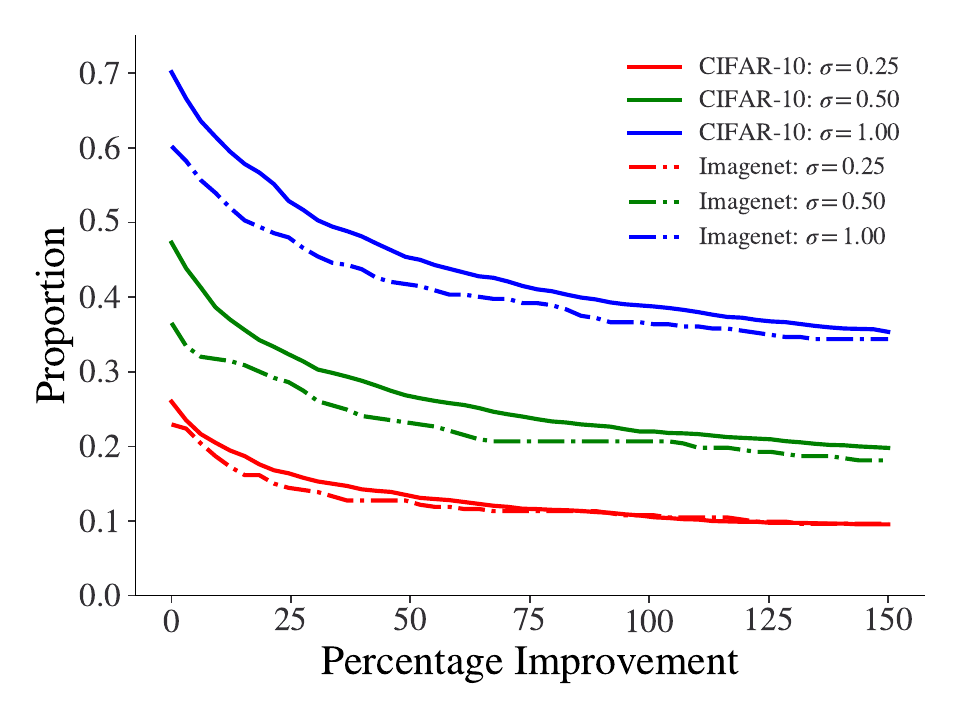}
    \end{center}
\caption{Proportion of multinomial samples for which there is a percentage improvement (of more than the value given in the horizontal-axis) in the ensemble relative to Cohen.}%
\label{fig:Differential_Performance}
\end{figure}

\section{Results}\label{sec:results}

To validate the applicability of our improved Gaussian mechanism and ensembling approach, experiments were performed for $\ell_2$-norm-bounded certifications utilising both CIFAR-$10$ \cite{krizhevsky2009learning}, and the latest face-blurred variant of ImageNet \cite{yang2021Imagenet}, which respectively exist under a MIT and a BSD $3-$Clause license. %
While this blurring has been shown to introduce a slight degradation of predictive performance, the privacy-preserving protections that are introduced are important for the integrity of vision research. Our results highlight ImageNet, due to presence of human indistinguishable adversarial examples within trained models~\cite{szegedy2013intriguing}.

To support this, training was performed upon NVIDIA P$100$ GPUs using in PyTorch \cite{NEURIPS2019_9015} and a cross-entropy loss, with a fixed random seed employed to ensure reproducibility. Certification employed $n=100,000$ samples for estimating expectations. Confidence intervals were set at $\alpha = 0.001$ for a $0.1 \%$ chance of any produced certification over-estimating the radius of certification. These parameters mirror those of Cohen~\etal \cite{cohen2019certified}. In the case of CIFAR-$10$, two GPUs and a $110$-layer residual network were trained for $90$ epochs under added noise $\sigma \in \{0.12, 0.25, 0.5, 1.0\}$, with certification then being applied using the appropriate matched $\sigma$. This training process used stochastic gradient descent subject to an initial learning rate of $0.1$, momentum of $0.9$ and weight decay of $10^{-4}$, and batch size set at $400$. The learning rate was refined in a step-wise fashion where $L_r = 0.1 \times 0.1^{\floor{\frac{e}{30}}}$, where $e$ is the current epoch. %

For ImageNet, training was performed using a mixed precision \cite{micikevicius2017mixed} ResNet-$50$ model using a ten-node system, where each node also had access to two GPUs. In order to understand the influence of noise on the more-complex model, training and certification were performed at levels $\sigma \in \{0.25, 0.5, 1, 2, 3, 4, 5\}$. Due to the increased complexity of ResNet-$50$, training was modified to match best practices under available system resources. Both GPU cores on every node were trained with a batch size of $200$, and in the fashion of \cite{goyal2017accurate} the learning rate is set at $L_r= 0.1563$, to be equivalent to a learning rate of $0.1$ for a single node training with a batch size of $256$. To improve convergence, the first three epochs were performed under $L_r = 0.01563$, before reverting to the original rate. This was then scaled by $0.1$ at the $30$, $60$ and $80$-th epochs. 


\subsection{Certified Accuracy} To assess the level of certified robustness provided, we adopt the now standard concept of \textit{certified accuracy}. This records the proportion of samples correctly predicted by randomised smoothing with a certified $r \geq R$, and in doing so captures both the accuracy of the model under noise, and the level of certification that can be provided to samples. Reflecting our analytic analysis of the softmax techniques, Figure~\ref{fig:Imagenet} demonstrates a uniform improvement over Lecuyer \etal, with both a larger certified radius as $r \to 0$, and a slower rate of decay in the certified accuracy as the radius increases. These performance increases reflect the analytic improvements demonstrated within Figure~\ref{fig:Theoretical_comparison_results}. 

While prior works have indicated that Cohen \etal \cite{cohen2019certified} uniformly outperformed all other techniques under a multinomial distribution \emph{by considering aggregate statistics}, our experiments clearly demonstrate that both our approach and Li \etal are able to certify a greater proportion of samples. This is a product of the implicit restriction in Cohen \etal that $E_0 \geq 0.5$. Relative to Li \etal, our technique yields improvements for samples in which $r < 1$, however beyond this point the distribution decays due to the tightness of the analytic bound as $E_0 \to 1$, relative to the other tested approaches. 

\begin{figure*} 
    \centering
  \subfloat[$\sigma=1.0$]{%
       \includegraphics[width=0.4\linewidth]{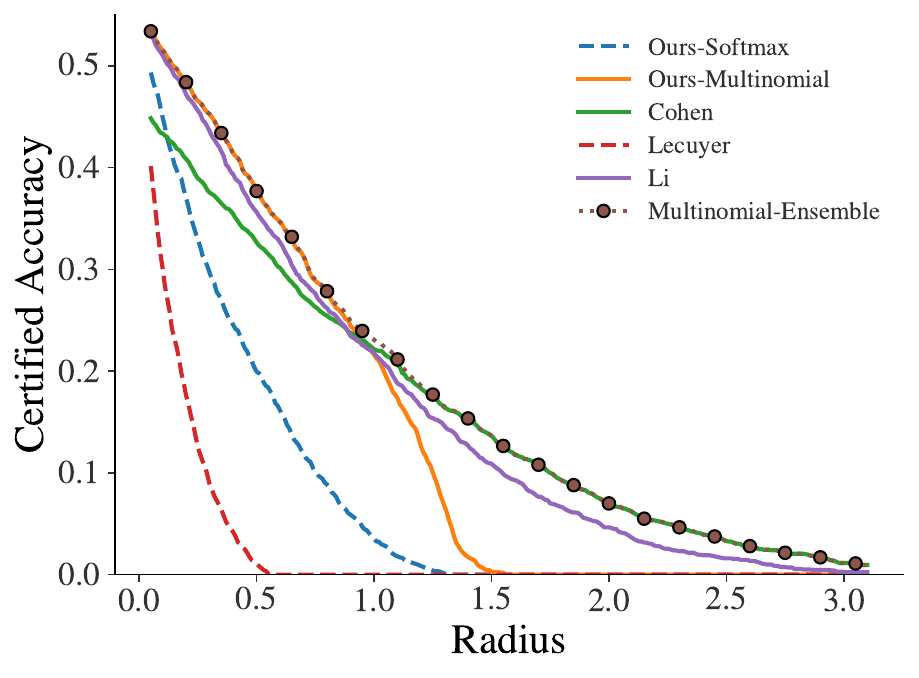}}
    \hfill
  \subfloat[Varying $\sigma$]{%
        \includegraphics[width=0.4\linewidth]{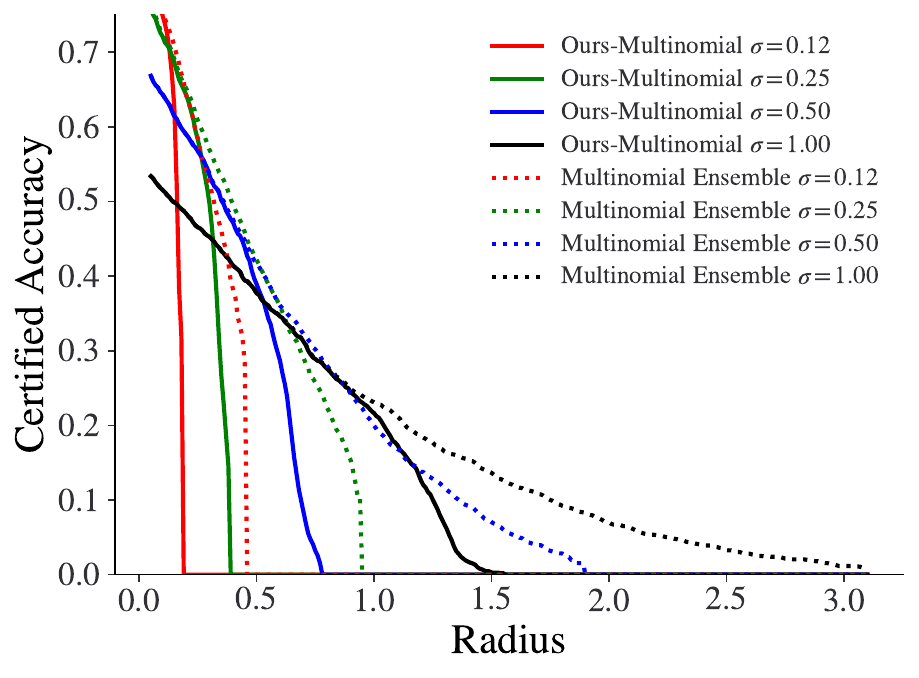}}
  \caption{CIFAR-$10$ Certified Accuracy, similar to Figure~\ref{fig:Imagenet}. Figure a) demonstrates the variable performance of each technique, while b) considers the performance of our technique against the Ensemble (that includes our approach).}%
\label{fig:Cifar}
\end{figure*}

The ensemble approach clearly improves upon both the number of samples certified, and the radii at which these samples are certified across all $\sigma$, as per Table~\ref{tab:large_table} and Figures~\ref{fig:Imagenet} and \ref{fig:Cifar}.  While the relative performance is maintained across both datasets, for CIFAR-$10$ there is an across the board increase in the overall certified accuracy, due to the decreased prediction difficulty in the $10$-class CIFAR relative to the $1000$-class ImageNet. Such differences in relative performance align with the multinomial comparison of Figure~\ref{fig:Theoretical_comparison_results} and underscore the importance of our proposed ensemble certification approach, as it leverages the regions of the simplex of output score in which each technique produces the largest certification radii.

Figure~\ref{fig:Differential_Performance} reinforces this, by demonstrating the proportion of samples for which the ensemble produces more than a given percentage-improvement, relative to Cohen \etal. We reiterate the observation of Algorithm~\ref{alg:Multinomial-C} abstaining from certifying a greater number of samples than the alternate techniques, and thus the ensemble produces an infinite percentage improvement relative to Cohen \etal for these samples. More broadly, the level of out-performance of the Ensemble relative to Cohen \etal is confirmed by Table~\ref{tab:results_tab}, which demonstrates the relative performance of the Ensemble against the prior state of the art of Cohen \etal. We note that as the Wilcoxon test produces highly significant statistics as it is comparing the prior state of the art in Cohen \etal to the ensemble, which also incorporates Cohen \etal. However, broadly Table~\ref{tab:results_tab} demonstrate that as $\sigma$ increases the Ensemble is able to reliably improve upon a significant proportion of samples, with a notable effect on the mean certification.

\begin{table*}
\caption{Metrics of performance, comparing the ensemble to the state of the art of Cohen \etal. Here the columns \textit{p} and \textit{Statistic} refer to the results of the Wilcoxon signed rank test (with the Pratt signed-rank zero procedure)~\cite{pratt1959remarks}; \textit{Proportion} is the percentage of samples which yield an improved certification due to the ensembling process, and the median and mean (exluding where $L_{\text{Cohen}} = 0$) columns cover the improvements. (in both absolute and percentage-values).}%
\label{tab:results_tab}
  \centering
\begin{tabular}{lllllllll}
\toprule
         &      &                 &                   &          & \multicolumn{2}{c}{Median}       & \multicolumn{2}{c}{Mean}               \\
    Dataset     &   $\sigma$   & p               & Statistic         & Proportion & Absolute & $\%$ & Absolute & $\%$ \\
\cmidrule(r){1-2} \cmidrule(r){3-4} \cmidrule(r){5-5} \cmidrule(r){6-7} \cmidrule(r){8-9}
         
Cifar    & $0.12$ &  $< 10^{-5}$ & $> 10^{4}$ & $14.4\%$   & $0$           & -              & $0.0020$      & $14.9\%$         \\
         & $0.25$ & $< 10^{-5}$ & $> 10^{4}$ & $26.1\%$   & $0$           & -              & $0.0079$      & $52.2\%$         \\
         & $0.5$  & $< 10^{-5}$ & $> 10^{4}$ & $47.4\%$   & $0$           & -              & $0.035$       & $46.1\%$         \\
         & $1.0$  & $< 10^{-5}$ & $> 10^{4}$ & $70.2\%$   & $0.079$       & $32.2\%$         & $0.12$        & $307.1\%$        \\
Imagenet & $0.25$ & $< 10^{-5}$ & $> 10^{4}$ & $22.9\%$   & $0$           & -              & $0.0083$      & $37.1\%$         \\
         & $0.5$  & $< 10^{-5}$ & $> 10^{4}$ & $36.5\%$   & $0$           & -              & $0.031$       & $181.3\%$        \\
         & $1.0$  & $< 10^{-5}$ & $> 10^{4}$ & $60.2\%$   & $0.053$       & $16.6\%$         & $0.12$        & $99.2\%$ \\       
 \bottomrule
\end{tabular}
\end{table*}

\subsection{Computational Costs} The nature of randomised smoothing---in that it requires repeated sampling---inherently introduces a significant computational cost, even if this process can be trivially parallelised. Intuitively it would appear that the computational cost of any ensembling approach would scale multiplicatively with the number of techniques being employed within the ensemble. However, as the expectations can be reused across all the ensembled techniques, this dominant component of the computational cost only has to be performed once. In practice, on NVIDIA P$100$ GPU the time to calculate the radius of certification (after the expectations have been estimated) is less than $0.1$ seconds for per technique, which is equivalent to less than the cost of $100$ performing samples under noise, as result that is borne out by Figure~\ref{fig:IMAGENET_Time}.

When considering  computational cost, it is important to emphasise that mechanistic improvements in how we perform certifications are useful not just for the increases in certified radius, but also to potentially decrease the computational cost. Due to the inherent link between the sample size and uncertainty levels, and these uncertainty levels with the certified radius (as is seen in Figure~\ref{fig:IMAGENET_Proportion}), \emph{improved certifications can be considered as an offset to the number of samples required, leading to commensurate decreases in the overall computational time}. %

\begin{figure}
    \begin{center}
        \includegraphics[width=0.4\textwidth]{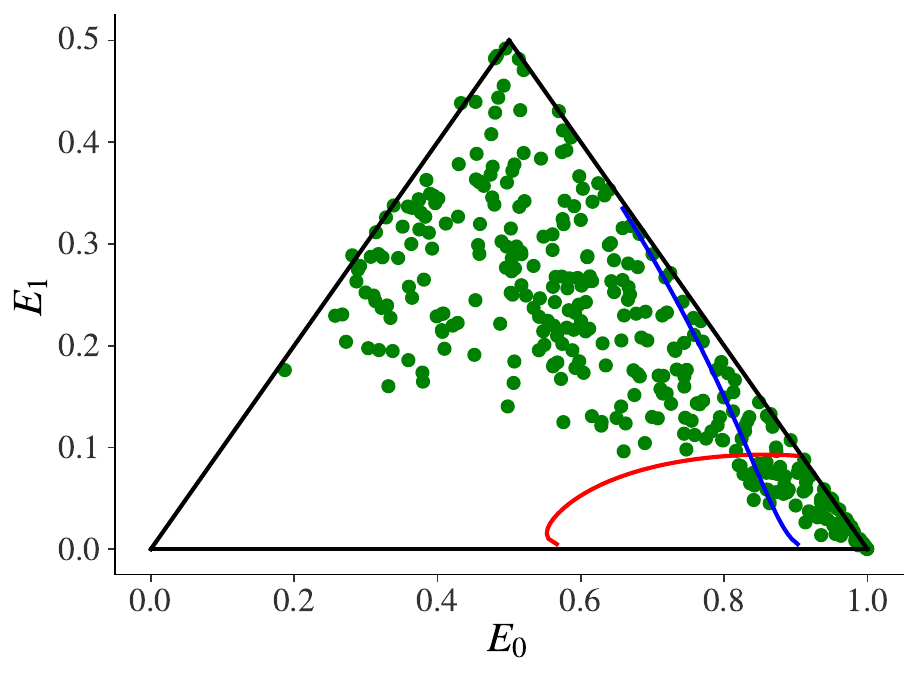}
    \end{center}
\caption{%
Distribution of multinomial output expectations when the samples are drawn from CIFAR-10, subject to the parameters $\sigma = 1$ and  $n = 100,000$, in a fashion equivalent to Figure~\ref{fig:Scatters}.%
}%
\label{fig:CIFAR_SCATTERS}
\end{figure}

\subsection{Simplex Coverage} As was established within Figure~\ref{fig:Theoretical_comparison_results} and Section~\ref{sec:related}, any experimental comparison of these approaches is an implicit function of the model, dataset, training procedure, and hyperparameters employed in training. This dependency is visible in Figures~\ref{fig:Scatters} and \ref{fig:CIFAR_SCATTERS}, in which transitioning from CIFAR-$10$ to ImageNet induces a \emph{distributional shift} in the distribution of label-space expectations towards the region of the simplex of output scores that favours our certification approach, with the average $E_0$ and $E_1$ both decreasing. That this shift occurs reflects the greater semantic similarities between classes within ImageNet, which results in output expectations that are more evenly spread across classes. 

This sensitivity to the input parameters is reinforced by Table~\ref{tab:large_table}, which demonstrates the inherent coupling between predictive difficulty (as indicated by the Top-1 accuracy) and a translation of the output distribution towards a region that is favourable to our differential privacy based approach. These changes in the output distribution in turn induce a shift in performance, from the initial strong performance of Cohen \etal (with exception to the proportion of samples certified with radii greater than $0.05$), to metrics that uniformly favour our new approach as $\sigma > 1$. These results also demonstrate that our approach can certify $9\%$ more samples above $0.05$ at $\sigma = 1$, and $49\%$ more as $\sigma$ is increased to $5$. Increasing $\sigma$ above $1$ also leads to our technique exhibiting a monotonic improvement in the median certified radii, which stands in stark contrast to the approach of Cohen \etal, which exhibits a consistent decrease in the median certification radius to $0$ at $\sigma = 5$. This behaviour in Cohen \etal is a product of the smoothing effect of additive noise, which results in fewer and fewer samples having a highest class expectation above $0.5$, preventing certification. 

Both Figure~\ref{fig:Scatters} and Table~\ref{tab:large_table} demonstrate that our analytic comparison and ensemble frameworks take advantage of simplex regions of differential performance to \emph{generate ensemble certifications that produce consistent, best-in-class results}. Moreover, we can be confident that the outperformance of this technique will be maintained irrespective of the complexity of the underlying dataset.

This work presents both ensembling and disaggregated analysis in the context of an $\ell_2$ bounded threat model, due to the relative maturity of attacks in this space.  However, our approaches are equally applicable to analysing certification performance against \emph{any} threat model, and future works expanding the oeuvre of certified threat models should exploit the techniques described within this work in order to better understand and maximise certification performance.

\subsection{Alternative Training Approaches}

\begin{figure*}[!ht]
    \centering
  \subfloat[$\sigma=1.0$]{%
       \includegraphics[width=0.38\linewidth]{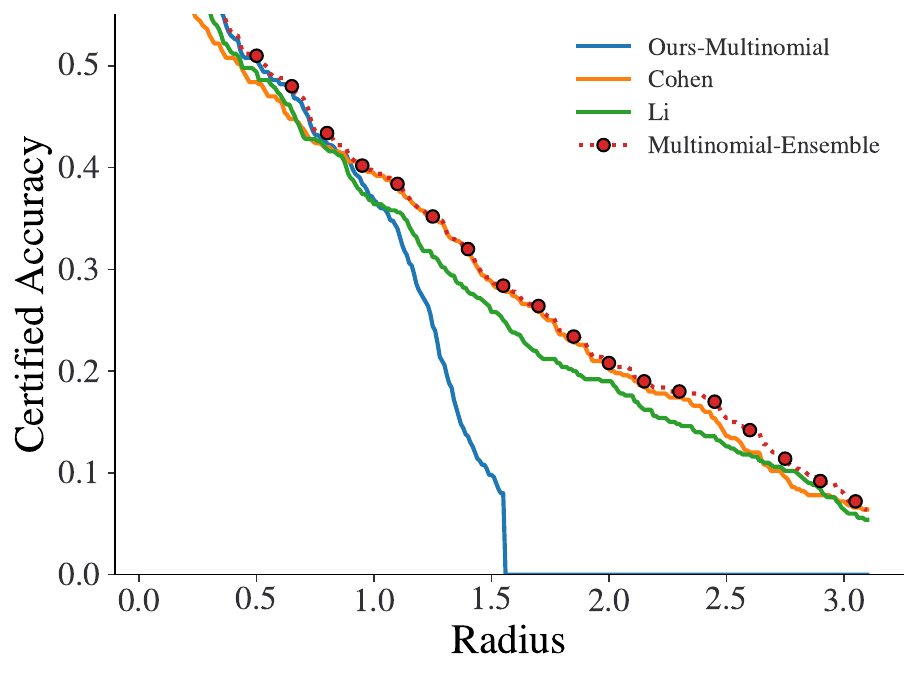}}
    \hfill
  \subfloat[Multinomial]{%
        \includegraphics[width=0.38\linewidth]{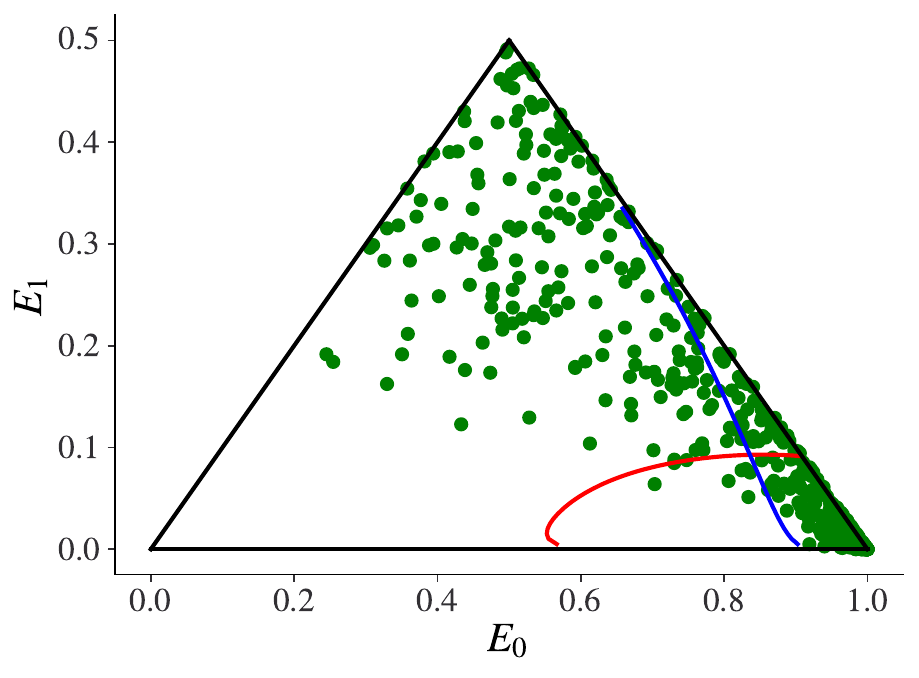}}
\caption{%
Certified Accuracy and distribution of Multinomial output expectations when the samples are drawn from CIFAR-10 when trained with MACER, subject to the parameters $\sigma = 1$ and  $n = 100,000$, in a fashion equivalent to Figure~\ref{fig:Scatters}.%
}%
\label{fig:MACER_Scatters}
\end{figure*}

The same independence of our approaches to a specific threat model also holds true when considering alternate certification frameworks--- including MACER~\cite{zhai2020macer}, denoising~\cite{carlini2022certified}, or Geometrically Informed Certified Robustness~\cite{cullen2022double}---as they each construct certifications in an identical manner. While these techniques shift the distributions seen within Figure~\ref{fig:CIFAR_SCATTERS}, our analysis still holds. While Figure~\ref{fig:MACER_Scatters} demonstrates that MACER shifts the overall point distribution towards the region where Cohen \etal is favoured, a significant proportion of samples are still located within the region where our new technique yields improved certifications. 

\section{Conclusions}

By considering certification performance through a disaggregated lens, this work demonstrates that it is possible to better understand the drivers of certification performance, from both a dataset and mechanistic perspective. This form of analysis demonstrated the utility of our multiple improvements to to the current oeuvre of certification techniques, including an improved differential privacy based Gaussian mechanism that for some samples can produce a more than two-fold increase in the achievable multinomial certification, or up to five-fold in the case of softmax certifications. These improvements are particularly evident in the case where the largest class expectation is diminished.

Beyond this, our work also demonstrates that a simple ensemble-of-certifications can reuse the costly components of certifications, in order to improve upon performance relative to any one single technique. This technique, which introduces almost no additional computational burden, is able to certify $98\%$ of samples above $r = 0.05$ at $\sigma = 1$ for ImageNet, relative to the $89\%$ achieved by the prior state-of-the-art in Cohen \etal \cite{cohen2019certified}. Our technique's advantage over other certification mechanisms grows with both the semantic complexity of the dataset and $\sigma$.

Through this works mechanisms, we have demonstrated how minor changes to certification systems can be used to construct larger certifications, which in turn would allow for a greater degree of confidence in the adversarial resistance of systems deployed in contexts where they may be implemented. Moreover, our approach of assessing certification performance within the context of the simplex of output scores has the potential to allow for a more nuanced view of adversarial risk. Operationalising this perspective, in the context of our improvements to the achievable radii of certification, has the potential to reduce reduce the need for domain experts to manually verify inputs that have the potential to be adversarially influenced, or to guide a greater understanding of adversarial risk in deployed systems.

\section*{Acknowledgements}

This research was undertaken using the LIEF HPC-GPGPU Facility hosted at the University of Melbourne. This Facility was established with the assistance of LIEF Grant LE170100200. This work was also  supported  in  part  by  the  Australian  Department  of  Defence  Next  Generation  Technologies  Fund, as part of the CSIRO/Data61 CRP AMLC project. Sarah Erfani is in part supported by Australian Research Council (ARC) Discovery Early Career Researcher Award (DECRA) DE220100680.
\section*{Resource Availability}

The full suite of code required to replicate the experiments contained within this work can be found at \url{https://github.com/andrew-cullen/ensemble-simplex-certifications}

\section*{Ethics Statement}

The techniques and processes described within this paper have the potential to decrease the vulnerability of deployed machine learning systems to adversarial example. However, in doing so there is also the potential to also counter beneficial applications of attacks, such as stylometric privacy. However, we believe that the value in minimising risks to deployed systems significantly outweighs these concerns.

\bibliographystyle{plain}
\bibliography{macros, main}

\end{document}